%% file: JELIA.tex
\documentclass[runningheads]{llncs}
\usepackage[utf8]{inputenc}
\usepackage{float}
\usepackage{graphicx}
\usepackage[colorinlistoftodos]{todonotes}
\usepackage{pdfpages}
\newcommand{\Ar}{\textit{Ar}}
\newcommand{\att}{\textit{att}}
\newcommand{\SCCs}{\textit{SCCs}}
\newcommand{\inn}{\texttt{in}}
\newcommand{\out}{\texttt{out}}
\newcommand{\undecided}{\texttt{undecided}}
\newcommand{\accept}{\texttt{accept}}
\newcommand{\reject}{\texttt{reject}}

\begin{document}
\title{Technical report of ``Empirical Study on Human\\ \textcolor{white}{.} \hspace{-13mm} Evaluation of Complex Argumentation Frameworks'' \hspace{-13mm} \textcolor{white}{.}}

\titlerunning{Empirical Study on Human Evaluation of Complex AFs}
%
\author{Marcos Cramer\inst{1}
\and
Mathieu Guillaume\inst{2}
}
%
%
\institute{International Center for Computational Logic, TU Dresden, Germany \\
\email{marcos.cramer@tu-dresden.de} \and
Centre for research in Cognitive Neuroscience (CRCN), Université Libre de Bruxelles,  Belgium\\
\email{maguilla@ulb.ac.be}}
\maketitle              
\begin{abstract}
In abstract argumentation, multiple argumentation semantics have been proposed that allow to select sets of jointly acceptable arguments from a given argumentation framework, i.e.\ based only on the attack relation between arguments. The existence of multiple argumentation semantics raises the question which of these semantics predicts best how humans evaluate arguments. Previous empirical cognitive studies that have tested how humans evaluate sets of arguments depending on the attack relation between them have been limited to a small set of very simple argumentation frameworks, so that some semantics studied in the literature could not be meaningfully distinguished by these studies. In this paper we report on an empirical cognitive study that overcomes these limitations by taking into consideration twelve argumentation frameworks of three to eight arguments each. These argumentation frameworks were mostly more complex than the argumentation frameworks considered in previous studies. All twelve argumentation framework were systematically instantiated with natural language arguments based on a certain fictional scenario, and participants were shown both the natural language arguments and a graphical depiction of the attack relation between them. Our data shows that grounded and CF2 semantics were the best predictors of human argument evaluation. A detailed analysis revealed that part of the participants chose a cognitively simpler strategy that is predicted very well by grounded semantics, while another part of the participants chose a cognitively more demanding strategy that is mostly predicted well by CF2 semantics.
\end{abstract}
\section{Introduction}
\input{intro}

\section{Preliminaries of Abstract Argumentation Theory}
\label{sec:prelim}
\input{prelim}

\section{Cognitive Variability of Humans}
\label{sec:variability}

Given that this paper presents findings of a cognitive empirical study to an audience whose scientific expertise lies mainly in areas outside of cognitive science, we present some general background from cognitive science that will help to make our methodological choices and our discussion of the results more understandable. 

Humans are heterogeneous by nature; they differ from each other with respect to 
their cognitive abilities~\cite{anastasi1958differential}.  Cronbach~\cite{Cronbach57} claimed that human heterogeneity is actually a major disturbance in the conduction of empirical studies.  Cognitive variability has thus been mostly considered as an undesirable random noise in cognitive studies.  This disturbance is even more problematic in the case of empirical studies that evaluate complex cognitive processes such as logical thinking and reasoning.  Indeed, the inherent difficulty of such tasks not only emphasizes human differences relative to pure cognitive abilities (such as intelligence), but also involves motivational aspects that are crucial to obtain a reliable performance from the participant~\cite{weiner1972theories}.  In order to test the cognitive plausibility of abstract argumentation theory by minimizing unwanted bias purely related to cognition and motivation properties, we set up a methodology that favored rational thinking during the assessment.  

Previous results showed that individual performance, which has generally been reported to be quite poor in pure logic and reasoning tasks, could actually be enhanced by cooperative discussion with peers. For instance, faced with the Wason selection task~\cite{wason1966reasoning}, humans solving the task in groups achieved a level of insight that was qualitatively superior to the one achieved by single individuals~\cite{Geil98,Augustinova08}. Additionally, and more generally, discussion with peers was shown to substantially improve motivation to solve a given task~\cite{piaget1995sociological}. For these reasons, we decided to incorporate in our methodology a cooperative discussion to help participants to elaborate and enrich their thinking. This collective step with peers was designed to obtain an evaluation of the justification status more reliable than a single individual judgment. Such reliability is crucial to test the cognitive plausibility of our predictions. 

\section{Design of the Study}
\label{sec:design}
Sixty-one undergraduate students participated in the empirical study (mean age = 20.8).  With the help of a questionnaire, we asked our participants to evaluate the acceptability status of natural language arguments.  The argument sets were set in the following fictional context: participants were located on an imaginary island, faced to conflicting information coming from various islanders, and they had to evaluate the arguments provided in order to hopefully find the location(s) of the buried treasure(s). We used such a fictional scenario to avoid as much as possible any unwanted interference from their general knowledge to make a decision about the acceptability of a given argument.

All the attacks between the arguments were based on information that a certain islander is not trustworthy. Consider for example the following set of arguments that we used in the study:
\begin{quote}
\textbf{Argument G:}  Islander Greg says that islander Hans is not trustworthy and that there is a treasure buried in front of the well. So we should not trust what Hans says, and we should dig up the sand in front of the well.\vspace{3mm}

\textbf{Argument H:}  Islander Hans says that islander Irina is not trustworthy and that there is a treasure buried behind the bridge. So we should not trust what Irina says, and we should dig up the sand behind the bridge.\vspace{3mm}

\textbf{Argument I:}  Islander Irina says that there is a treasure buried near the northern tip of the island. So we should dig up the sand near the northern tip of the island.\vspace{3mm}

\textbf{Argument J:}  Islander Jenny says that there is a treasure buried near the southern tip of the island. So we should dig up the sand near the southern tip of the island.
\end{quote}
Here argument G attacks argument H, because argument H is based on information from islander Hans, and argument G states that islander Hans is not trustworthy. Similarly, argument H attacks argument I, whereas arguments I and J do not attack any argument because they do not state that someone is not trustworthy. (Note that participants were informed that there might be multiple treasures, so there is no conflict between a treasure being in one place and a treasure being in another.)

\begin{figure}
\vspace{-2mm}
\begin{center}
\includegraphics[width=0.8\textwidth]{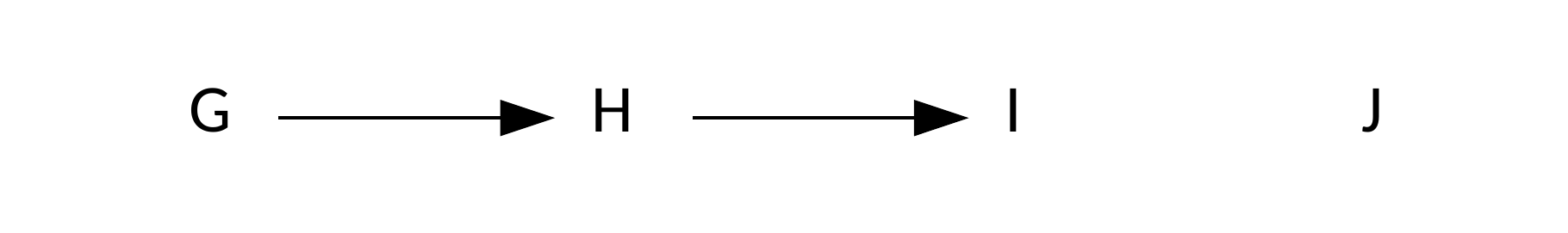}
\vspace{-6mm}
\end{center}
 \caption{Graphical visualization of the AF corresponding to the natural language arguments G, H, I and J.}
 \label{fig:GHIJ}
\vspace{-1mm}
\end{figure}

As the natural language arguments where quite long and complex, we presented to the participants not only the natural language arguments, but also a graphical visualization of the corresponding AF. For example, Figure \ref{fig:GHIJ} depicts the AF corresponding to the natural language argument set presented above.

Before the start of the questionnaire, we showed to the participants examples of three simple AFs of two arguments each, namely a unilateral attack from an argument to another one, a bilateral attack between two arguments, and two arguments without any attack between them. These examples were presented both as sets of natural language arguments and as graphically depicted AFs, and the correspondence between these two modes of presentation were explained.

Participants were instructed to make a judgment about each argument by ticking a box labeled \emph{accept}, \emph{undecided} or \emph{reject}. For the purpose of making these judgments, participants were explicitly instructed to follow the principle that they trust an islander as long as they do not have a good reason to believe that this islander is not trustworthy. We explained these three possible judgments on the six arguments from the three simple AFs that we showed as examples. Note that on these simple AFs, the justification status of each argument is the same in each of the six semantics considered in this paper, so that our explanations about these examples did not prime the participants to favor one of the semantics over the others.

Similarly as in our previous study~\cite{cramer2018empirical}, our methodology incorporated a group discussion to stimulate more rational thinking: Participants had to first respond individually to each argument from an argument set, then in a second step they had to collaboratively discuss with their peers about the set under consideration, and finally they had to make a final individual judgment.  We formed twenty groups of mostly three participants each (exceptionally two or four participants).  The questionnaire had two versions, each consisting in six different AFs, for a total of twelve argument frameworks. The full set of the argument sets used in our study can be found in the appendix. 

\section{Results and Discussion}
\label{sec:results}
\begin{figure}[H]
\vspace{-1mm}
  \begin{center}
  \begin{minipage}{0.48\textwidth}
  \begin{center}
    \includegraphics[width=1.02\textwidth]{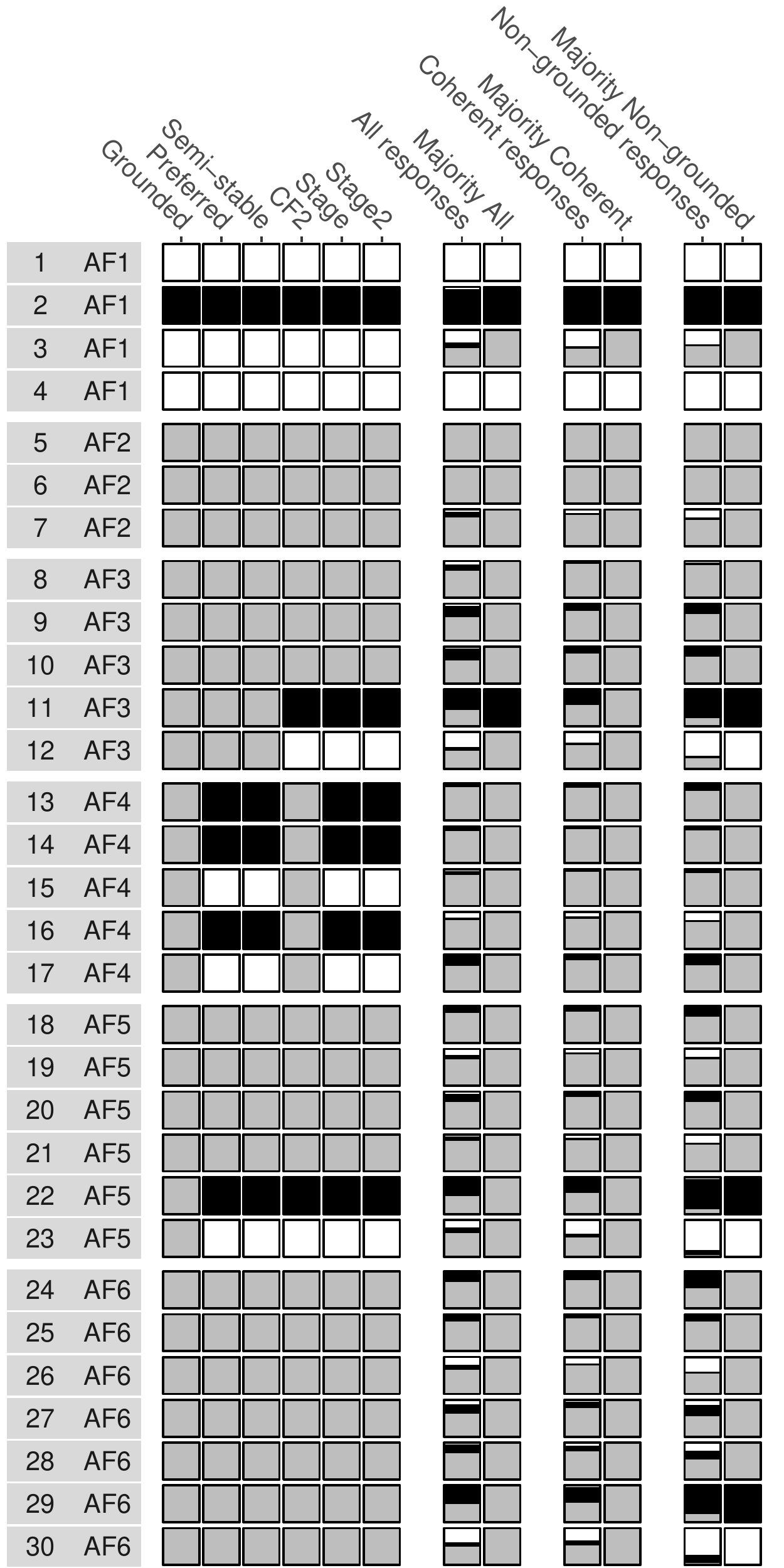}
  \end{center}
  \end{minipage}
  \hfill
  \begin{minipage}{0.48\textwidth}
  \begin{center}
    \includegraphics[width=1.02\textwidth]{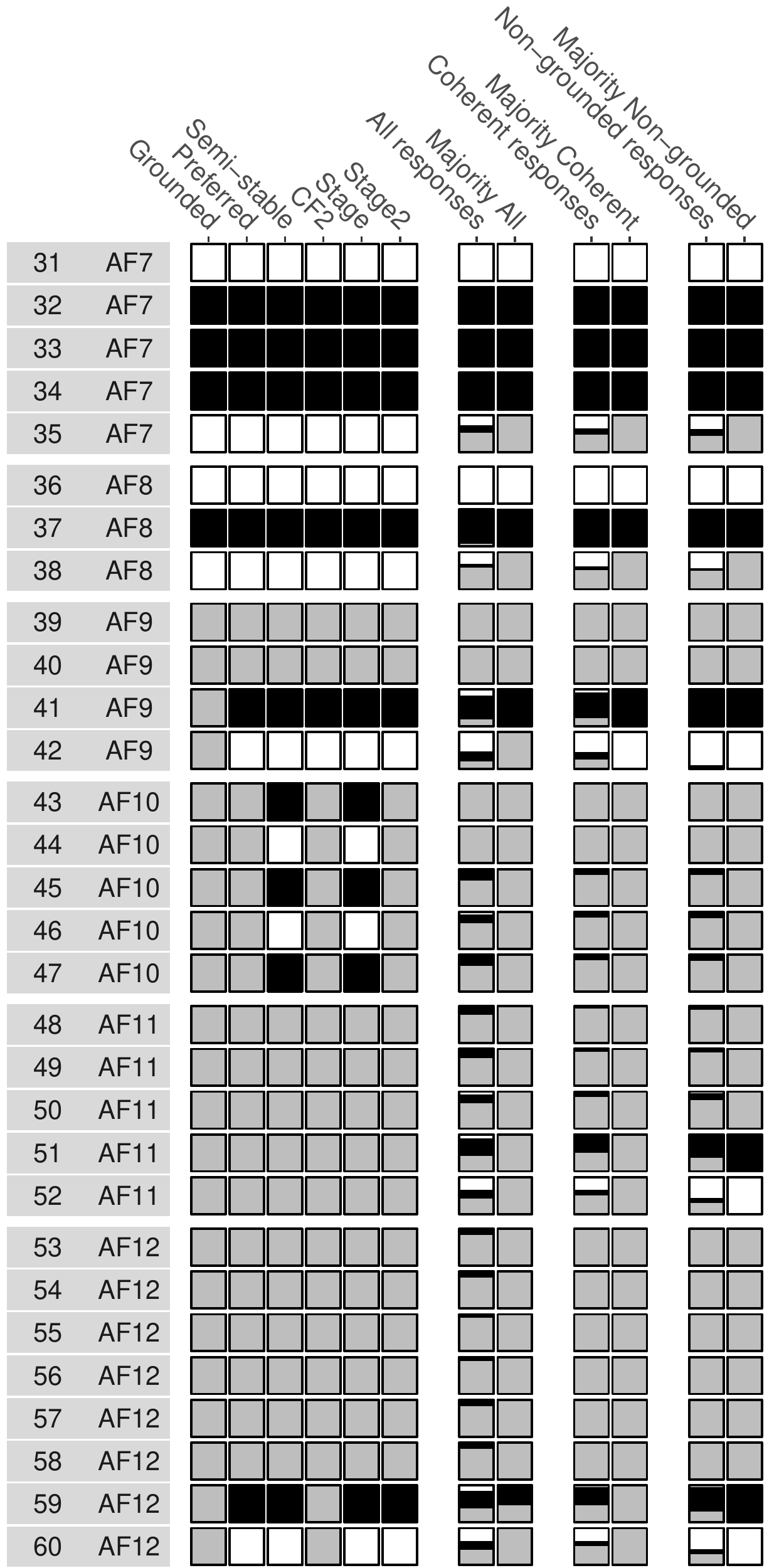}
  \end{center}
  \end{minipage}
  \end{center}
  \vspace{-2mm}
\caption{Visualization of the predictions and of the results. Each line represents one of the 60 arguments in our study. The squares represent theoretical predictions according to the six semantics and as well as final individual responses (average response and majority response) in three categories of participants: all participants, coherent participants and coherent non-grounded participants. White, black and gray stand for \emph{accept}, \emph{reject} and \emph{undecided} respectively. For representing the proportion of different responses, the corresponding square has been filled with white, gray and black area in proportion to the number of \emph{accept}, \emph{reject} and \emph{undecided} judgments made by the participants.}
\label{fig:results}
\end{figure}

Figure \ref{fig:results} summarizes both the theoretical predictions and the final individual response of participants in our study. In the first six columns of the figure, we explicitly represent the justification status of each argument with respect to each of the six semantics considered in this paper. We depict the justification status \emph{strongly accepted} as a white square, \emph{strongly rejected} as a black square, and \emph{undecided} as a gray square. In the next two columns, we have depicted the proportion of different responses made by the participants as well as the majority response. With the exception of argument 59, the arguments had a unique majority response, i.e.\ a response chosen more often than each of the other two responses, which is depicted by one of the three pure colors defined above. In the case of argument 59, \emph{reject} and \emph{undecided} were equally frequent responses, so the majority response square is depicted hald black and half gray.

In a first analysis, we assessed which semantics was the best to predict human evaluation of the acceptability of our natural language arguments. We computed the percentage of agreement between the predictions of each semantics and the final responses made by all participants. Predictions according to grounded semantics were correct in 74.97\%, preferred in 68.42\%, semi-stable in 62.84\%, CF2 in 75.46\%, stage in 62.79\%, and stage2 in 68.36\% of the cases. Exact binomial tests revealed that for all semantics, the proportion of correct predictions were significantly larger than the chance level (i.e., 33\%), all $p$s $< .001$. It is noteworthy that, in many cases, the semantics make the same prediction, so to evaluate the significance of the difference between any two semantics, we should not consider the general predictive accuracy, but rather focus on the instances where the two semantics under consideration differed. We thus conducted exact binomial tests between each pair of semantics, restricting to the arguments where different predictions were provided, and we observed that both grounded and CF2 were systematically better than the other semantics, all $p$s $< .001$. However, grounded and CF2 did not significantly differ from each other, $p=.212$. In other words, across all our participants, grounded and CF2 semantics were the semantics providing the best predictions. 

In order to get a better picture of the cognitive strategies employed by participants to evaluate arguments, we made some additional analysis of the data. We observed that participants mostly responded in a way that is coherent in the sense defined at the end of Section \ref{sec:prelim}. More precisely, 86.7\% of the responses were coherent, and 49 of the 61 participants (i.e.\ 80.3\% of the participants)
\linebreak

\noindent 
had more than 80\% coherent responses. Recall that the notion of coherence was based on properties that are satisfied in all semantics considered in this paper, 
so these results show that participants were mostly able to use cognitive strategies that are in line with these semantics-independent properties. We hypothesize that those 12 participants who had more than 20\% incoherent responses either did non fully understand the task or are outliers with respect to the cognitive strategy they employed. As we were interested in understanding the cognitive strategies employed by the majority of participants, we decided to disregard these 12 participants in the further analysis of our data. We use the expression \emph{coherent participants} to refer to the 49 participants that had more than 80\% coherent responses. The average and majority responses of coherent participants are depicted in Figure \ref{fig:results} in the two columns that are to the right of the columns depicting the responses of all participants.

Within coherent participants, predictions according to grounded semantics were correct in 82.79\%, preferred in 75.17\%, semi-stable in 68.10\%, CF2 in 82.24\%, stage in 67.14\%, and stage2 in 74.22\% of the cases. A paired $t$-test revealed that the predictions here were significantly better than the predictions across all participants, $t(5) = 12.38, p < .001$. This is in line with our hypothesis that the identified and excluded 12 participants were outliers. Once again, grounded and CF2 were the two best semantics, as confirmed by exact binomial tests restricting to the arguments for which the predictions differed, relevant $p$s $< .001$, and they did not significantly differ from each other, $p=.187$. Subsequently, within coherent participants, and similarly to our findings within all participants, grounded and CF2 were the best semantics to predict human responses.

Furthermore, we would like to point out that in the grounded semantics, 48 of the 60 arguments in our study were undecided. For this reason, the general strategy of choosing \emph{undecided} whenever there is some reason for doubt was a cognitively simple way to get full or almost full agreement with the prediction of grounded semantics. While it is an interesting observation that a significant number of participants chose this strategy for the task in our study, we were also interested in understanding better the cognitive strategy of those who did not make use of this simplifying general strategy. In order to get some insights about this cognitive strategy, we decided to make some additional analysis of our data restricted to those coherent participants that did not employ this grounded-leaning general strategy. For this purpose, we had to define a criterion for deciding who counts as not having applied the grounded-leaning general strategy. We chose to use the following criterion: If a participant made at least one coherent response that was not the response predicted by the grounded semantics, we considered this participant a \emph{non-grounded participant}. Of the 49 coherent participants, 27 were non-grounded participants according to this criterion, while 22 participants were \emph{grounded participants}. The average and majority responses of coherent non-grounded participants are depicted in the two last columns of Figure \ref{fig:results}.

Within coherent non-grounded participants, predictions according to grounded semantics were correct in 73.09\%, preferred in 73.70\%, semi-stable in 65.80\%, CF2 in 79.75\%, stage in 67.04\%, and stage2 in 74.94\% of the cases. In this case, CF2 alone was the best predictor in comparison to every other semantics, with the largest $p =.001$. 
This result provides further insights about the cognitive strategies adopted by participants: While grounded and CF2 semantics both provide adequate predictions of the human evaluation of the acceptability of the arguments, this is actually due to heterogeneous behavior from our participants. Our results suggest that 27 non-grounded participants used a more demanding cognitive strategy well predicted by CF2 whereas the other 22 grounded participants used a more straightforward strategy well predicted by grounded semantics.

\begin{figure}
\vspace{-2mm}
\begin{center}
\includegraphics[width=0.8\textwidth]{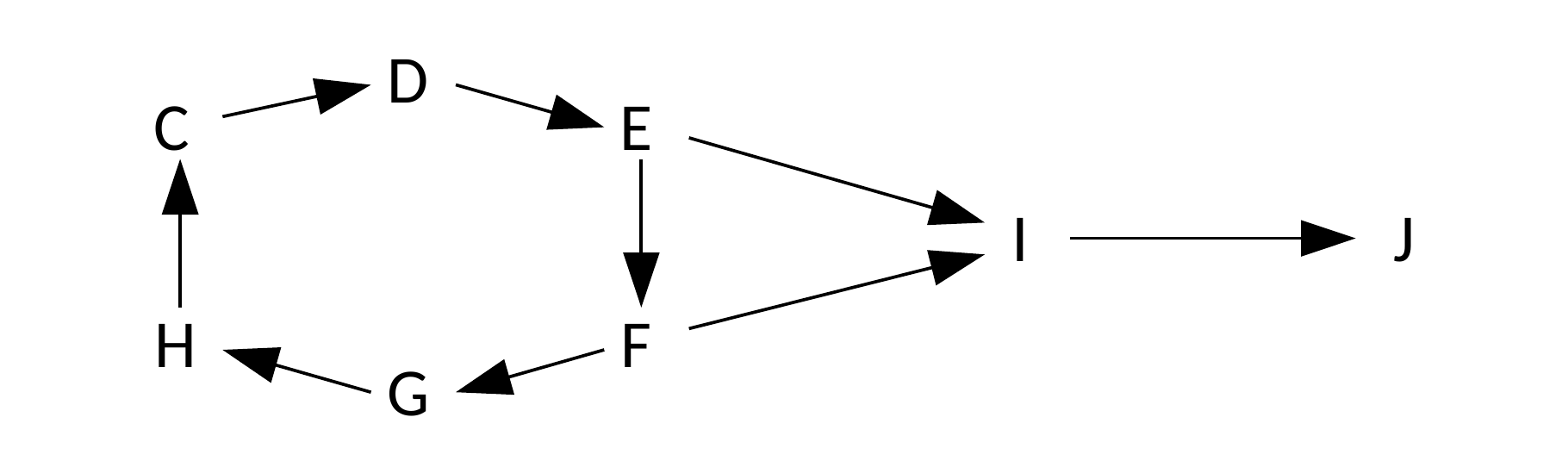}
\end{center}
\vspace{-6mm}
 \caption{AF in which other semantics made better prediction than CF2}
 \label{fig:CDEFGHIJ}
\vspace{-1mm}
\end{figure}

We would like to point out that the only two arguments in which some semantics other than CF2 predicted the judgments of coherent non-grounded participants better than CF2 were arguments 59 and 60 according to the numbering used in Figure \ref{fig:results}, which were arguments I and J in the AF depicted in Figure \ref{fig:CDEFGHIJ}. While in CF2 and grounded semantics both of these arguments are weakly undecided, in preferred, semi-stable, stage and stage2 semantics, I is strongly rejected and J is strongly accepted. 

Note that this AF contains a six-cycle, and the behavior of CF2 on a six-cycle was criticized by Dvo\v{r}\'ak~and~Gaggl~\cite{Dvorak16} as unintuitive and used as a motivation for introducing stage2 semantics. We included this AF in our study to test whether this criticism on CF2 semantics is in line with human judgments on such AFs, and our data does indeed support this criticism on CF2. However, all other arguments on which the predictions of CF2 and stage2 differed were judged by most coherent non-grounded participants more in line with CF2 than in line with stage2, so our data does not support stage2 as a good alternative to CF2. 

%
%
%

Taken together, this suggests that for the goal of predicting well human argument acceptance, it might be necessary to develop a novel semantics that behaves similarly to CF2 on most argumentation frameworks considered in our study but which treats even cycles of length 6 or more in the way they are treated by preferred, semi-stable, stage and stage2 semantics rather than in the way they are treated by CF2 semantics.
 
\section{Related work}
\label{sec:related}
While there have been multiple empirical studies that have evaluated the correspondence between human reasoning and formalism from abstract, structured and dialogical argumentation (see for example~\cite{cerutti2014formal,rosenfeld2016providing,hunter2017empirical,Polberg18}), only two previous studies have aimed at comparing human evaluation of arguments to abstract argumentation semantics: Rahwan~\textit{et~al.}~\cite{rahwan2010behavioral} tested how humans evaluate two simple argumentation frameworks, the \emph{simple reinstatement} framework with three arguments and the \emph{floating reinstatement} framework with four arguments. In a recent paper~\cite{cramer2018empirical}, the authors of the present study have improved Rahwan~\textit{et~al.}'s methodology and applied this improved methodology to three different argumentation frameworks, namely the two AFs already considered by Rahwan as well as the \emph{3-cycle reinstatement} framework with five arguments.

Since the simple reinstatement framework is treated in the same way by all standard semantics, in Rahwan~\textit{et~al.}'s study only the floating reinstatement framework was able to distinguish between different semantics. While this allowed Rahwan~\textit{et~al.} to conclude that for the floating reinstatement argumentation frameworks the preferred semantics predicts human evaluation of arguments better than the grounded semantics, it did not allow to distinguish preferred semantics from other non-grounded semantics like semi-stable, stage or CF2. By including the 3-cycle reinstatement framework in our previous study, we were able to observe that naive-based semantics like CF2, stage or stage2 are better predictors for human argument evaluation than admissibility-based semantics like preferred or semi-stable (see~\cite{cramer2018empirical}). However, the AFs used in that study still did not allow to distinguish between the different naive-based semantics, nor did they allow to distinguish preferred from semi-stable semantics. The present study was designed to overcome this limitation.

We now compare the results from the present paper with those from our recent paper~\cite{cramer2018empirical}. The current study confirmed the result of the previous study that CF2 semantics is a better predictor for human argument evaluation than preferred semantics, and extended this result by also showing that CF2 is a better predictor than semi-stable, stage, stage2 semantics. The previous study had additionally suggested that both preferred and CF2 semantics are better predictors than grounded semantics, whereas the current study suggests that grounded semantics is as good a predictor as CF2 semantics. We believe that the main reason for this apparent mismatch lies in the fact that our present study used more complex argumentation frameworks and instantiated them with a fictional scenario, which made the reasoning task cognitively more challenging and therefore led to more participants making use of the simplifying strategy of choosing \emph{undecided} whenever there is some reason for doubt.

Both Rahwan~\textit{et~al.}'s study and our previous study made use of natural language arguments that referred to real-world entities and actions rather than to a purely fictional scenario as in the present study. While this reference to real-world entities and actions reduces the cognitive load for participants, it also allows them to make use of their world knowledge in judging the arguments. But as the goal of these studies was to predict argument evaluation based on the attack relation between arguments rather than based on the content of the argument and the world knowledge of the participants, this interference of world knowledge was undesirable. By making use of a fictional scenario in the present study we avoided this undesirable feature of the previous studies.

\section{Conclusion and Future Work}
\label{sec:conclusion}

In this paper we have reported on an empirical cognitive study in which we tested how humans judge the acceptability of arguments in complex argumentation frameworks. A detailed analysis of our results revealed that part of the participants chose a cognitively simpler strategy that is predicted very well by grounded semantics, while another part of the participants chose a cognitively more demanding strategy that is mostly predicted well by CF2 semantics.

The present study suggests multiple paths for future research. As for future research within formal argumentation, our study suggests that researchers in this field who are interested in developing formalisms that correspond well to human argumentation should direct their attention more to CF2 and similarly-behaved semantics. More precisely, given that the cognitively more demanding strategy was predicted well by CF2 semantics with the exception of the AF involving a six-cycle, it seems worthwhile to develop and study novel semantics that behave similarly to CF2 on most argumentation frameworks considered in our study but which treat even cycles of length 6 or more in the way they are treated by preferred, semi-stable, stage and stage2 semantics rather than in the way they are treated by CF2 semantics. Furthermore, given that in the context of structured argumentation frameworks like ASPIC+ (see \cite{modgil2014aspic+}) the rationality postulate of \emph{Closure under Strict Rules} is violated for not admissibility-based semantics like CF2, further research is required to find a method to satisfy this rationality postulate in structured argumentation while using an argumentation semantics that corresponds well to human judgments on argument acceptability.

As for future empirical work related to the work presented in this paper, it would be good to empirically test whether our tentative explanation that we have given in Section \ref{sec:related} for explaining the mismatch between the current study and our previous study (see \cite{cramer2018empirical}) is correct. Furthermore, it would be good if some future empirical study could overcome a limitation that all existing empirical studies on abstract argumentation theory have, namely the limitation that they can only compare the semantics on the single-outcome justification status, thus ignoring some of the information present in the full set of extensions provided by each semantics. For overcoming this limitation, a novel approach to designing empirical cognitive studies for testing argumentation semantics needs to be developed.

\bibliographystyle{myplain}

\input{JELIA.bbl}


\section*{Supplementary material}

## Appendix

This appendix provides the instructions, the examples, and the argumentation frameworks that were used in the study presented in this paper. There are 12 argumentation frameworks (AFs) for a total of 60 arguments. The argumentation frameworks are instantiated with natural language arguments, which are presented together with their graphical visualizations, just like in the questionnaires used in the study. The natural language arguments are labeled with a letter, which is used in the graphical visualizations and was given in the questionnaire as well, and additionally with the number (indicated in parenthesis) to which we refer in Figure 2.

**Instructions**

This questionnaire asks you to evaluate whether given arguments are to be accepted or to be rejected in a certain context. The context of all the arguments in this questionnaire is the following scenario, which we ask you to imagine:

You arrive at an island that is said to have multiple hidden treasures that you want to find. Luckily, many of the islanders know about where the treasures are hidden, and some of them are willing to help you. However, some of the islanders are trying to deceive you by giving you wrong information. You only have a limited amount of time for digging in search for treasures, so you want to make sure you make a good decision about where it is worth to look for a treasure and where it is not worth it.

The only way you can possibly find out about which islanders are trustworthy and which ones are not is by asking each islander about the trustworthiness of the other islanders. You decide to follow the principle that you trust an islander as long as you don't have a good reason to believe that this islander is not trustworthy.

First you go around asking islanders for information, and you take notes of which islander told you what. Furthermore, based on what the islanders tell you, you write down arguments for searching a treasure in certain places and for not trusting certain islanders. Once you have asked all islanders, you look back at the list of arguments that you have formulated.

Now your task is to decide which of these arguments you accept and which ones you reject. If there is no way to decide between accepting and rejecting an argument, you can also mark it as undecided. Please tick one box per argument (i.e. per line).

In order to help you understand the logical relations between the arguments, we also provide a graphical representation of the logical attacks between arguments. This is explained in more detail in the examples in the following sections.

**Example 1**

> **Argument A**: Islander Amy says that there is a treasure buried between the two tall palm trees. So we should dig up the sand between the two palm trees.
>
> **Argument B**: Islander Berta says that islander Amy is not trustworthy and that there is a treasure buried behind the tower. So we should not trust what Amy says, and we should dig up the sand behind the tower.

Here argument B provides a reason to not trust Amy, and thus to reject argument A. So argument B can be used as a counterargument against argument A. Another way of saying this is to say that argument B attacks argument A.

Argument A, on the other hand, does not provide a reason to reject argument B, so argument A cannot be used as a counterargument against argument B. In this case we say that argument A does not attack argument B.

The logical relation between argument A and argument B can be depicted as follows:

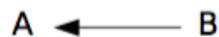

The arrow from B to A means that argument B attacks argument A. Since argument A does not attack argument B, there is no arrow back from A to B.

Given the principle that you trust an islander as long as you don't have a good reason to believe that this islander is not trustworthy, you should accept argument B, since it is based on trusting islander Berta and you don't have any reasons to believe that Berta is not trustworthy. And given that you accept argument B, you have to reject argument A, since argument A is based on trusting islander Amy and argument B provides a good reason to believe that Amy is not trustworthy.

One can state the reasoning in the previous paragraph more simply as follows: Since no argument attacks argument B, argument B should be accepted. Since an argument that is accepted attacks argument A, argument A should be rejected.

To indicate the choice of rejecting argument A and accepting argument B, you should tick the boxes as follows:

|  | Accept | Undecided | Reject |
|---|---|---|---|
| Argument A | ○ | ○ | ⦿ |
| Argument B | ⦿ | ○ | ○ |

**Example 2**

> **Argument C**: Islander Claire says that islander Daniel is not trustworthy and that there is a treasure buried on the shore of the lake. So we should not trust what Daniel says, and we should dig up the sand on the shore of the lake.
>
> **Argument D**: Islander Daniel says that islander Claire is not trustworthy and that there is a treasure buried next to the large rock. So we should not trust what Claire says, and we should dig up the sand next to the large rock.

Here argument C provides a reason to not trust Daniel, and thus to reject argument D. So argument C attacks argument D. Similarly, argument D provides a reason to reject argument C, i.e. argument D also attacks argument C. So graphically, the situation is as follows:

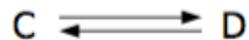

Since the two arguments are in conflict, we cannot accept both of them. But there is no reason to prefer one argument over the other, so we cannot decide which one to accept and which one to reject. For this reason, we consider both arguments as undecided.

To indicate that you consider both arguments undecided, you should tick the boxes as follows:

| | Accept | Undecided | Reject |
|---|---|---|---|
| Argument C | ○ | ◉ | ○ |
| Argument D | ○ | ◉ | ○ |

**Example 3**

> **Argument E**: Islander Ernst says that there is a treasure buried between the two hills. So we should dig up the sand between the two hills.
>
> **Argument F**: Islander Fiona says that there is a treasure buried behind the big rock. So we should dig up the sand behind the big rock.

There is no conflict between the information provided by the two arguments. In other words, neither argument attacks the other. This can be depicted graphically as follows:

$$E \qquad F$$

Here, there is no arrow between argument E and argument F, as neither of them attacks the other.

Since no reason is given to believe that either islander Ernst or islander Fred is not trustworthy, the principle that you trust an islander as long as you don't have a good reason to believe that this islander is not trustworthy implies that you should trust both Ernst and Fiona. Thus you should accept both arguments.

One can state this reasoning more simply as follows: Since no argument is attacked by any argument, we can accept both arguments.

To indicate that you accept both arguments, you should tick the boxes as follows:

|  | Accept | Undecided | Reject |
|---|---|---|---|
| Argument E | ● | ○ | ○ |
| Argument F | ● | ○ | ○ |

## Argumentation Framework 1

**Argument G (1)**: Islander Greg says that islander Hans is not trustworthy and that there is a treasure buried in front of the well. So we should not trust what Hans says, and we should dig up the sand in front of the well.

**Argument H (2)**: Islander Hans says that islander Irina is not trustworthy and that there is a treasure buried behind the bridge. So we should not trust what Irina says, and we should dig up the sand behind the bridge.

**Argument I (3)**: Islander Irina says that there is a treasure buried near the northern tip of the island. So we should dig up the sand near the northern tip of the island.

**Argument J (4)**: Islander Jenny says that there is a treasure buried near the southern tip of the island. So we should dig up the sand near the southern tip of the island.

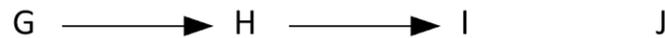

## Argumentation Framework 2

**Argument T (5)**: Islander Tina says that islander Umberto is not trustworthy and that there is a treasure buried between the olive trees. So we should not trust what Umberto says, and we should dig up the sand between the olive trees.

**Argument U (6)**: Islander Umberto says that islander Tina and islander Victor are not trustworthy and that there is a treasure buried to the south of the swamp. So we should not trust what Tina and Victor say, and we should dig up the sand to the south of the swamp.

**Argument V (7)**: Islander Victor says that there is a treasure buried to the north of the swamp. So we should dig up the sand to the north of the swamp.

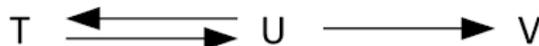

## Argumentation Framework 3

**Argument L (8)**: Islander Lisa says that islander Mila and islander Olivia are not trustworthy and that there is a treasure buried on the peak of the mountain. So we should not trust what Mila and Olivia say, and we should dig up the sand on the peak of the mountain.

**Argument M (9)**: Islander Mila says that islander Neil and islander Olivia are not trustworthy and that there is a treasure buried next to the old wall. So we should not trust what Neil and Olivia say, and we should dig up the sand next to the old wall.

**Argument N (10)**: Islander Neil says that islander Lisa and islander Olivia are not trustworthy and that there is a treasure buried between the two oak trees. So we should not trust what Lisa and Olivia say, and we should dig up the sand between the two oak trees.

**Argument O (11)**: Islander Olivia says that islander Peter is not trustworthy, and that there is a treasure buried near the eastern tip of the island. So we should not trust what Peter says, and we should dig up the sand near the eastern tip of the island.

**Argument P (12)**: Islander Peter says that there is a treasure buried near the southern tip of the island. So we should dig up the sand near the southern tip of the island.

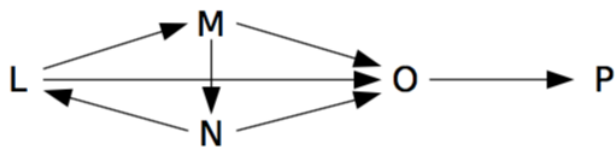

# Argumentation Framework 4

**Argument R (13)**: Islander Ron says that islander Sarah is not trustworthy and that there is a treasure buried next to the harbor. So we should not trust what Sarah says, and we should dig up the sand next to the harbor.

**Argument S (14)**: Islander Sarah says that islander Tom is not trustworthy and that there is a treasure buried north of the forest. So we should not trust what Tom says, and we should dig up the sand north of the forest.

**Argument T (15)**: Islander Tom says that islander Ron, islander Sarah and islander Ursula are not trustworthy and that there is a treasure buried south of the forest. So we should not trust what Ron, Sarah and Ursula say, and we should dig up the sand south of the forest.

**Argument U (16)**: Islander Ursula says that islander Vincent is not trustworthy, and that there is a treasure buried next to the old monument. So we should not trust what Vincent says, and we should dig up the sand next to the old monument.

**Argument V (17)**: Islander Vincent says that there is a treasure buried next to the ruins. So we should dig up the sand next to the ruins.

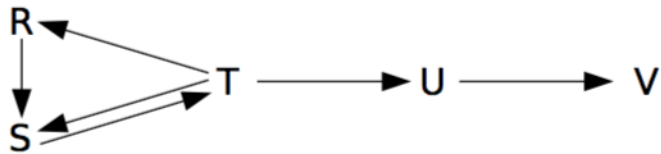

## Argumentation Framework 5

**Argument W (18)**: Islander Walter says that islander Xavier is not trustworthy and that there is a treasure buried between the ponds. So we should not trust what Xavier says, and we should dig up the sand between the ponds.

**Argument X (19)**: Islander Xavier says that islander Yanis and islander Anna are not trustworthy and that there is a treasure buried next to the temple. So we should not trust what Yanis and Anna say, and we should dig up the sand next to the temple.

**Argument Y (20)**: Islander Yanis says that islander Zoe and islander Anna are not trustworthy and that there is a treasure buried to the east of the lake. So we should not trust what Zoe and Anna say, and we should dig up the sand to the east of the lake.

**Argument Z (21)**: Islander Zoe says that islander Walter is not trustworthy, and that there is a treasure buried between the two highest mountains. So we should not trust what Walter says, and we should dig up the sand between the two highest mountains.

**Argument A (22)**: Islander Anna says that islander Bella is not trustworthy, and that there is a treasure buried to the west of the lake. So we should not trust what Bella says, and we should dig up the sand to the west of the lake.

**Argument B (23)**: Islander Bella says that there is a treasure buried to the north of the lake. So we should dig up the sand to the north of the lake.

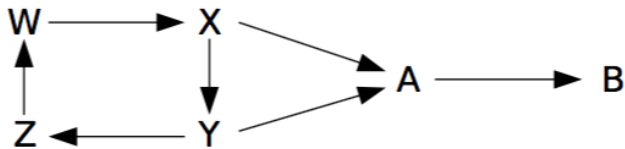

# Argumentation Framework 6

**Argument C (24)**: Islander Charlie says that islander Dorothy is not trustworthy and that there is a treasure buried to the south of the high mountain. So we should not trust what Dorothy says, and we should dig up the sand to the south of the high mountain.

**Argument D (25)**: Islander Dorothy says that islander Emma is not trustworthy and that there is a treasure buried to the west of the high mountain. So we should not trust what Emma says, and we should dig up the sand to the west of the high mountain.

**Argument E (26)**: Islander Emma says that islander Fred and islander Hannah are not trustworthy and that there is a treasure buried to the east of the high mountain. So we should not trust what Fred and Hannah say, and we should dig up the sand to the east of the high mountain.

**Argument F (27)**: Islander Fred says that islander George and islander Hannah are not trustworthy, and that there is a treasure buried to the north of the high mountain. So we should not trust what George and Hannah say, and we should dig up the sand to the north of the high mountain.

**Argument G (28)**: Islander George says that islander Charlie is not trustworthy, and that there is a treasure buried to the west of the village. So we should not trust what Charlie says, and we should dig up the sand to the west of the village.

**Argument H (29)**: Islander Hannah says that islander Ivan is not trustworthy, and that there is a treasure buried to the south of the village. So we should not trust what Ivan says, and we should dig up the sand to the south of the village.

**Argument I (30)**: Islander Ivan says that there is a treasure buried to the north of the village. So we should dig up the sand to the north of the village.

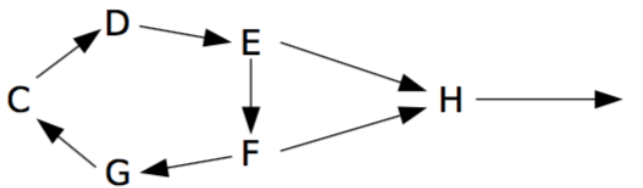

# Argumentation Framework 7

**Argument G (31)**: Islander Greg says that islander Hans, islander Irina and islander Jenny are not trustworthy and that there is a treasure buried in front of the well. So we should not trust what Hans, Irina and Jenny say, and we should dig up the sand in front of the well.

**Argument H (32)**: Islander Hans says that islander Ken is not trustworthy and that there is a treasure buried behind the bridge. So we should not trust what Ken says, and we should dig up the sand behind the bridge.

**Argument I (33)**: Islander Irina says that islander Ken is not trustworthy and that there is a treasure buried near the northern tip of the island. So we should not trust what Ken says, and we should dig up the sand near the northern tip of the island.

**Argument J (34)**: Islander Jenny says that islander Ken is not trustworthy and that there is a treasure buried near the southern tip of the island. So we should not trust what Ken says, and we should dig up the sand near the southern tip of the island.

**Argument K (35)**: Islander Ken says that there is a treasure buried near the mouth of the river. So we should dig up the sand near the mouth of the river.

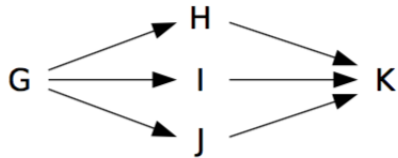

## Argumentation Framework 8

**Argument T (36)**: Islander Tina says that islander Umberto is not trustworthy and that there is a treasure buried between the olive trees. So we should not trust what Umberto says, and we should dig up the sand between the olive trees.

**Argument U (37)**: Islander Umberto says that islander Victor is not trustworthy and that there is a treasure buried to the south of the swamp. So we should not trust what Victor says, and we should dig up the sand to the south of the swamp.

**Argument V (38)**: Islander Victor says that islander Umberto is not trustworthy and that there is a treasure buried to the north of the swamp. So we should not trust what Umberto says, and we should dig up the sand to the north of the swamp.

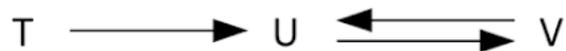

## Argumentation Framework 9

**Argument L (39)**: Islander Lisa says that islander Mila and islander Neil are not trustworthy and that there is a treasure buried on the peak of the mountain. So we should not trust what Mila and Neil say, and we should dig up the sand on the peak of the mountain.

**Argument M (40)**: Islander Mila says that islander Lisa and islander Neil are not trustworthy and that there is a treasure buried next to the old wall. So we should not trust what Lisa and Neil say, and we should dig up the sand next to the old wall.

**Argument N (41)**: Islander Neil says that islander Olivia is not trustworthy and that there is a treasure buried between the two oak trees. So we should not trust what Olivia says, and we should dig up the sand between the two oak trees.

**Argument O (42)**: Islander Olivia says that there is a treasure buried near the eastern tip of the island. So we should dig up the sand near the eastern tip of the island.

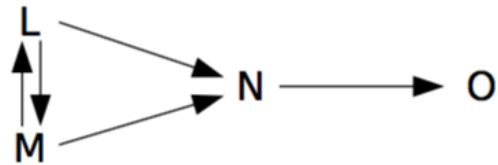

# Argumentation Framework 10

**Argument R (43)**: Islander Ron says that islander Sarah is not trustworthy and that there is a treasure buried next to the harbor. So we should not trust what Sarah says, and we should dig up the sand next to the harbor.

**Argument S (44)**: Islander Sarah says that islander Ron and islander Tom are not trustworthy and that there is a treasure buried north of the forest. So we should not trust what Ron and Tom say, and we should dig up the sand north of the forest.

**Argument T (45)**: Islander Tom says that islander Ursula is not trustworthy and that there is a treasure buried south of the forest. So we should not trust what Ursula says, and we should dig up the sand south of the forest.

**Argument U (46)**: Islander Ursula says that islander Vincent is not trustworthy, and that there is a treasure buried next to the old monument. So we should not trust what Vincent says, and we should dig up the sand next to the old monument.

**Argument V (47)**: Islander Vincent says that islander Tom is not trustworthy and that there is a treasure buried next to the ruins. So we should not trust what Tom says, and we should dig up the sand next to the ruins.

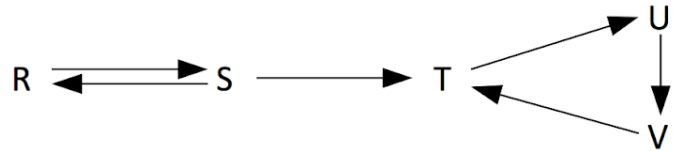

# Argumentation Framework 11

**Argument W (48)**: Islander Walter says that islander Xavier is not trustworthy and that there is a treasure buried between the ponds. So we should not trust what Xavier says, and we should dig up the sand between the ponds.

**Argument X (49)**: Islander Xavier says that islander Yanis and islander Zoe are not trustworthy and that there is a treasure buried next to the temple. So we should not trust what Yanis and Zoe say, and we should dig up the sand next to the temple.

**Argument Y (50)**: Islander Yanis says that islander Walter and islander Zoe are not trustworthy and that there is a treasure buried to the east of the lake. So we should not trust what Walter and Zoe say, and we should dig up the sand to the east of the lake.

**Argument Z (51)**: Islander Zoe says that islander Anna is not trustworthy, and that there is a treasure buried between the two highest mountains. So we should not trust what Anna says, and we should dig up the sand between the two highest mountains.

**Argument A (52)**: Islander Anna says that there is a treasure buried to the west of the lake. So we should dig up the sand to the west of the lake.

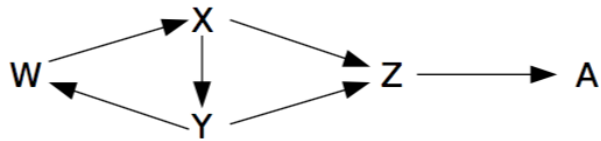

# Argumentation Framework 12

**Argument C (53)**: Islander Charlie says that islander Dorothy is not trustworthy and that there is a treasure buried to the south of the high mountain. So we should not trust what Dorothy says, and we should dig up the sand to the south of the high mountain.

31

**Argument D (54)**: Islander Dorothy says that islander Emma is not trustworthy and that there is a treasure buried to the west of the high mountain. So we should not trust what Emma says, and we should dig up the sand to the west of the high mountain.

**Argument E (55)**: Islander Emma says that islander Fred and islander Ivan are not trustworthy and that there is a treasure buried to the east of the high mountain. So we should not trust what Fred and Ivan say, and we should dig up the sand to the east of the high mountain.

**Argument F (56)**: Islander Fred says that islander George and islander Ivan are not trustworthy, and that there is a treasure buried to the north of the high mountain. So we should not trust what George and Ivan say, and we should dig up the sand to the north of the high mountain.

**Argument G (57)**: Islander George says that islander Hannah is not trustworthy, and that there is a treasure buried to the west of the village. So we should not trust what Hannah says, and we should dig up the sand to the west of the village.

**Argument H (58)**: Islander Hannah says that islander Charlie is not trustworthy, and that there is a treasure buried to the east of the village. So we should not trust what Charlie says, and we should dig up the sand to the east of the village.

**Argument I (59)**: Islander Ivan says that islander John is not trustworthy, and that there is a treasure buried to the south of the village. So we should not trust what John says, and we should dig up the sand to the south of the village.

**Argument J (60)**: Islander John says that there is a treasure buried to the north of the village. So we should dig up the sand to the north of the village.

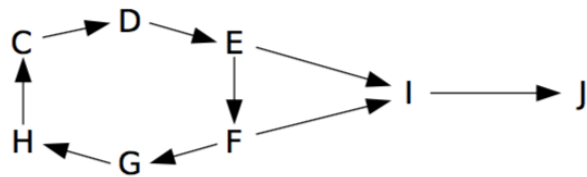



\end{document}

%% file: intro.tex
The formal study of argumentation is an important field of research within AI~\cite{rahwan2009argumentation}. 
One important methodological approach in the formal study of argumentation is abstract argumentation as introduced by Dung~\cite{dung1995acceptability}, in which one models arguments by abstracting away from their internal structure to focus on the relation of attacks between them, i.e.\ on the relation between a counterargument and the argument that it counters. 
Multiple \emph{argumentation semantics} have been proposed in the literature as criteria for selecting acceptable arguments based on the structure of the attack relation between the arguments (see~\cite{Baroni18}). Given that the applicability of abstract argumentation theory to human reasoning is desirable, this situation gives rise to the question which semantics best predicts the judgments that humans make about the acceptability of arguments based on the attack relation between the arguments.

There have been two previous empirical cognitive studies that have tested how humans evaluate sets of arguments depending on the attack relation between them, namely a 2010 study by Rahwan~et~al.~\cite{rahwan2010behavioral} as well as a recent study by the authors of this paper~\cite{cramer2018empirical}. These previous studies have been limited to small sets of very simple argumentation frameworks, so that some semantics studied in the literature could not be meaningfully distinguished by these studies. The study presented in this paper was designed to overcome this limitation by taking into account a larger number of argumentation frameworks, including some that are more complex than any of the argumentation frameworks used in previous studies. 

When studying human evaluation of argumentation frameworks, it is important to fill the arguments with meaning rather than just presenting abstract graphs to humans, as most humans will not be able to properly understand the reasoning task if it is presented in a purely abstract manner (see Chapter 4 of \cite{evans1993human}).  For this reason, we instantiated the argumentation frameworks with natural language arguments, as was also done by the two previous studies cited above. When instantiating argumentation frameworks with natural language arguments, one needs to be careful in choosing the natural language arguments in such a way that 
for each pair of arguments, humans judge the existence and directionality of the attack between the two arguments as intended by the designers of the study. In a recent paper~\cite{cramer2018directionality}, we have presented the results of two empirical cognitive studies that tested how humans judge the existence and directionality of attacks between pairs of arguments. 
Note that designing sets of natural language arguments that -- based on our findings in this recent paper -- correspond to complex argumentation frameworks is a highly non-trivial task. 

In order to approach this task in a systematic way, we carefully designed a fictional scenario in which information from multiple sources is analyzed, and developed a method to instantiate argumentation frameworks of arbitrary complexity with natural language arguments related to this fictional scenario. All attacks between arguments were based on undercutting the trustworthiness of a a source, as our recent paper suggests that undercutting the trustworthiness of a source corresponds well to a unidirectional attack~\cite{cramer2018directionality}. We used this method to design twelve sets of natural language arguments corresponding to twelve argumentation frameworks that had been carefully chosen to highlight the differences between existing argumentation semantics. As the natural language arguments were quite long and complex, we presented to the participants not only the natural language arguments, but also a graphical visualization of the corresponding argumentation framework.

We compared the results of our study to six widely studied argumentation semantics, namely to \emph{grounded}, \emph{preferred}, \emph{semi-stable}, \emph{CF2}, \emph{stage} and \emph{stage2} semantics. More precisely, we compare them to a three-valued justification status that can be defined with respect to each semantics. Due to certain considerations about these justification statuses, we do not separately consider \emph{complete} and \emph{stable} semantics in this paper.

The results of our study show that grounded and CF2 semantics were the best predictors of human argument evaluation. A detailed analysis revealed that part of the participants chose a cognitively simpler strategy that is predicted very well by grounded semantics, while another part of the participants chose a cognitively more demanding strategy that is mostly predicted well by CF2 semantics. In the discussion of our results, we pay special attention to the observation that the only argumentation framework for which CF2 semantics predicted the outcome of this cognitively more demanding strategy not as well as some other semantics was a framework including a six-cycle.

The rest of this paper is structured as follows: In Section \ref{sec:prelim}, we present the formal preliminaries of abstract argumentation theory that are required in this paper. In particular, we define stage, CF2 and stage2 semantics and the three justification statuses used in this paper. In Section \ref{sec:variability}, we present some general background from cognitive science that will help to make our methodological choices and our discussion of the results more understandable. The design of our study is explained in Section \ref{sec:design}. In Section \ref{sec:results}, we present and discuss the results of our study. Section \ref{sec:conclusion} concludes the paper and suggests directions for future research.

%% file: prelim.tex
We will assume that the reader is familiar with the basics of abstract argumentation theory as introduced by Dung~\cite{dung1995acceptability} and as explained in its current state-of-the-art form by Baroni~et~al.~\cite{Baroni18}. In particular, we will assume that the reader knows the notion of an \emph{argumentation framework} (AF) as well as the \emph{complete}, \emph{grounded}, \emph{stable}, \emph{preferred} and \emph{semi-stable} argumentation semantics, both in their traditional extension-based variant and in their label-based variant~\cite{baroni2011introduction,Baroni18}. In this section we furthermore define stage, CF2 and stage2 semantics as well as the notions of \emph{strong acceptance} and \emph{strong rejection}.

Stage semantics was first defined by Verheij~\cite{verheij1996two}. The idea behind it is that we minimize the set of arguments that are not accepted despite not being attacked by an accepted argument. To formalize this, we first need some auxiliary notions:

\begin{definition}
Let $F = \langle \Ar,\att \rangle$ be an AF and let $S \subset \Ar$ be a set of arguments. We say that $S$ is \textit{conflict-free} iff there are no arguments $b,c \in S$ such that $b$ attacks $c$ (i.e. such that $(b,c) \in \att$). We define $S^+ := \{b \in \Ar \mid \textit{for some } a \in S,$ $(a,b) \in \att \}$
\end{definition}

Now stage extensions are defined as follows:

\begin{definition}
Let $F = \langle \Ar,\att \rangle$ be an AF and let $S \subset \Ar$. Then $S$ is a stage extension of $F$ iff $S$ is a conflict-free set such that $S \cup S^+$ is
maximal with respect to set inclusion.
\end{definition}

CF2 semantics was first introduced by Baroni~\textit{et~al.}~\cite{baroni2005scc}. The idea behind it is that we partition the AF into \emph{strongly connected components} and recursively evaluate it component by component by choosing maximal conflict-free sets in each component and removing arguments attacked by chosen arguments. We formally define it following the notation of Dvo\v{r}\'ak~and~Gaggl~\cite{Dvorak16}. For this we first need some auxiliary notions:

\begin{definition}
Let $F = \langle \Ar,\att \rangle$ be an AF. We define $a \sim b$ iff either $a=b$ or there is an \att-pat path from $a$ to $b$ and there is an \att-path from $b$ to $a$. The equivalence classes under the equivalence relation $\sim$ are called \emph{strongly connected components} (SCCs) of $F$. We denote the set of SCCs of $F$ by $\SCCs(F)$. Given $S \subseteq \Ar$, we define $D_F(S) := \{b \in \Ar \mid \exists a \in S: (a,b) \in \att \land a \not \sim b \}$. 
\end{definition}

We now recursively define CF2 extensions as follows:

\begin{definition}
 Let $F = \langle \Ar,\att \rangle$ be an AF, and let $S \subseteq \Ar$. Then $S$ is a CF2 extension of $F$ iff either
 \vspace{-2mm}
 \begin{itemize}
  \item $|\SCCs(F)| = 1$ and $S$ is a maximal conflict-free subset of $A$, or
  \item $|\SCCs(F)| > 1$ and for each $C \in \SCCs(F)$, $S \cap C$ is a CF2 extension of $F|_{C - D_F(S)}$.
 \end{itemize}
\end{definition}

Stage2 semantics as introduced by Dvo\v{r}\'ak~and~Gaggl~\cite{dvorak2012incorporating,Dvorak16} combines features of stage and CF2 semantics by making use of the SCC-recursive scheme as in the definition of CF2, but using stage semantics rather than maximal conflict-freeness as the criterion to apply within a single SCC:

\begin{definition}
 Let $F = \langle \Ar,\att \rangle$ be an AF, and let $S \subseteq \Ar$. Then $S$ is a stage2 extension of $F$ iff either
 \vspace{-2mm}
 \begin{itemize}
  \item $|\SCCs(F)| = 1$ and $S$ is a stage extension of $A$, or
  \item $|\SCCs(F)| > 1$ and for each $C \in \SCCs(F)$, $S \cap C$ is a stage2 extension of $F|_{C - D_F(S)}$.
 \end{itemize}
\end{definition}

While the grounded extension of an AF is always unique, an AF with cycles may have multiple preferred, semi-stable, CF2, stage and stage2 extensions. In our experiment, however, participants were asked to make a single judgment about each argument, so we compare their judgments to the \emph{justification status} of arguments according to various semantics (see~\cite{wu2010labelling,Baroni18}), as the justification status is always unique for each argument. In particular, we focus on the justification statuses \emph{strongly accepted}, \emph{strongly rejected} and \emph{weakly undecided}, which can be defined as follows:

\begin{definition}
  Let $F = \langle \Ar,\att \rangle$ be an AF, let $\sigma$ be an argumentation semantics, and let $a \in A$ be an argument. We say that $a$ is \emph{strongly accepted with respect to} $\sigma$ iff for every $\sigma$-extension $E$ of $F$, $a \in E$. We say that $a$ is \emph{strongly rejected with respect to} $\sigma$ iff for every $\sigma$-extension $E$ of $F$, some $b \in E$ attacks $a$. We say that $a$ is \emph{weakly undecided} iff it is neither strongly accepted nor strongly rejected.
\end{definition}

Note that in the labeling approach, strong acceptance of $a$ corresponds to $a$ being labeled $\inn$ by all labelings, strong rejection of $a$ corresponds to $a$ being labeled $\out$ by all labelings, and a weakly undecided status for $a$ of corresponds to $a$ either being labeled $\undecided$ by at least one labeling, or $a$ being labeled $\inn$ by some labeling and $\out$ by some other labeling.

When comparing semantics to responses by humans, we will use these three justification statuses as a predictor for the human judgments to \emph{accept} an argument, \emph{reject} it or consider it \emph{undecided}. 

For some argumentation frameworks, stable semantics does not provide any extension whatsoever, which leads to the rather unintuitive situation that all arguments are both strongly accepted and strongly rejected. For this reason, we do not consider stable semantics as a potential predictor for human argument evaluation in this paper. The justification status with respect to complete semantics is always identical to the justification status with respect to grounded semantics, so that for the rest of the paper we do not separately consider complete semantics. 


We would like to point our three properties that the justification status of an argument $a$ satisfies in all semantics considered in this paper:
\begin{itemize}
 \item If all arguments attacking $a$ are strongly rejected, then $a$ is strongly accepted.
 \item If some argument attacking $a$ is strongly accepted, then $a$ is strongly rejected.
 \item If not all arguments attacking $a$ are strongly rejected, then $a$ is not strongly accepted.
\end{itemize}

We use this observation to define a notion of \emph{coherence} of a human judgment of the status of an argument with respect to the judgments of the other arguments in the same framework.

\begin{definition}
 Let $F = \langle \Ar,\att \rangle$ be an AF, and let $j: \Ar \rightarrow \{\accept,\reject,$ $\undecided\}$ be a function that represents three-valued judgments on the arguments in $\Ar$. Given an argument $a \in \Ar$, we say that the judgment of $j$ on $a$ is \emph{coherent} iff the following three properties are satisfied:
 \begin{itemize}
 \item If $j(b) = \reject$ for each argument $b$ that attacks $a$, then $j(a) = \accept$.
 \item If $j(b) = \accept$ for some argument $b$ that attacks $a$, then $j(a) = \reject$.
 \item If $j(b) = \undecided$ for some argument $b$ that attacks $a$, then either $j(a) = \undecided$ or $j(a) = \reject$.
 \end{itemize}
\end{definition}